\documentclass{article}

\usepackage{arxiv}
\usepackage[utf8]{inputenc} 
\usepackage[T1]{fontenc}    
\usepackage[hyphens]{url}   
\usepackage{hyperref}       
\usepackage{booktabs}       
\usepackage{amsfonts}       
\usepackage{amsmath}       %
\usepackage{nicefrac}       
\usepackage{microtype}      
\usepackage{lipsum}		
\usepackage{graphicx}
\usepackage{tabularx}
\usepackage{subcaption}
\usepackage{doi}
\usepackage{gensymb}
\usepackage{authblk}
\usepackage{multicol}
\usepackage{float}
\usepackage{caption}
\usepackage{changepage}
\usepackage{enumitem}
\usepackage[usenames,dvipsnames]{xcolor}

\title{Fluid Viscosity Prediction Leveraging Computer Vision and Robot Interaction}
\date{} 					

\author{\large\textbf{Jong Hoon Park}}
\author{\textbf{Gauri Pramod Dalwankar}}
\author{\textbf{Alison Bartsch}}
\author{\textbf{Abraham George}}
\author{\textbf{Amir Barati Farimani}}

\affil{\large Carnegie Mellon University}
\affil{\small \texttt{\{jonghoop, gdalwank, abartsch, aigeorge\}@andrew.cmu.edu, barati@cmu.edu}}

\hypersetup{
pdftitle={Fluid Viscosity Prediction Leveraging Computer Vision and Robot Interaction},
pdfsubject={cs.LG},
pdfauthor={J. H. Park, G. P. Dalwankar, A. Bartsch, A. George, A. B. Farimani}
}

\begin{document}

\maketitle

\begin{multicols}{2} 
\begin{abstract}
\begin{adjustwidth}{-1cm}{-1cm}
    Accurately determining fluid viscosity is crucial for various industrial and scientific applications. Traditional methods of viscosity measurement, though reliable, often require manual intervention and cannot easily adapt to real-time monitoring. With advancements in machine learning and computer vision, this work explores the feasibility of predicting fluid viscosity by analyzing fluid oscillations captured in video data. The pipeline employs a 3D convolutional autoencoder pretrained in a self-supervised manner to extract and learn features from semantic segmentation masks of oscillating fluids. Then, the latent representations of the input data, produced from the pretrained autoencoder, is processed with a distinct inference head to infer either the fluid category (classification) or the fluid viscosity (regression) in a time-resolved manner. 
    When the latent representations generated by the pretrained autoencoder are used for classification, the system achieves a 97.1\% accuracy across a total of 4,140 test datapoints. Similarly, for regression tasks, employing an additional fully-connected network as a regression head allows the pipeline to achieve a mean absolute error of 0.258 over 4,416 test datapoints. This study represents an innovative contribution to both fluid characterization and the evolving landscape of Artificial Intelligence, demonstrating the potential of deep learning in achieving near real-time viscosity estimation and addressing practical challenges in fluid dynamics through the analysis of video data capturing oscillating fluid dynamics.
\end{adjustwidth}
\end{abstract}



\section{Introduction}
\label{sec:intro}
    Viscosity is a fundamental property of fluids that plays an important role in numerous industrial applications such as product quality control \cite{oil_pred,difficult_viscosity,viscosity-with-Yolo}, engineering pump design \cite{pump}, and food processing \cite{foodsci}. Conventionally, its measurement is achieved using viscometers or rheometers, which require a labor-intensive process. The experimenter must select an appropriate device based on the fluid under examination, and the measurement can be time-consuming depending on fluid type, temperature, and viscometer size \cite{difficult_viscosity}. To address those experimental challenges, there have been many recent approaches to exploit the automation capability of robots or vision-based systems in both industries and academic laboratories, which shows great potential to significantly alleviate the labor-intensive processes and time cost \cite{robotchem1,robotchem2,robotchem3,robotchem4,robotchem5}.
    
    Inferring fluid properties like viscosity from visual data is a challenging task due to the complex nature of fluids, both in behavior and detection. Nevertheless, the ability to infer their viscosities directly from visual information is highly valuable for autonomous fluid handling systems as cameras are readily available and the inference process can occur in near real-time.
    Achieving real-time viscosity measurement offers significant practical benefits, particularly in settings where controlled conditions for image acquisition, such as consistent lighting and precise filming angles, can be maintained. In industrial processes or laboratory environments where such controls are feasible, this method could greatly enhance efficiency by providing instantaneous and non-invasive viscosity assessments, thereby streamlining quality control and monitoring procedures. Moreover, its application in automated systems could facilitate rapid decision-making in fluid handling and characterization, leading to increased productivity and accuracy in operations.
    
    In the context of perceiving viscosities of different fluids, humans can intuitively \textit{perceive} viscosities by observing the oscillation dynamics of fluids. Each fluid type exhibits a unique damping coefficient discernible in its oscillation pattern, and the decay rate offers a direct correlation to the viscosity of the fluid \cite{Tactile_sensing,viscous_decay}. For example, shaking a bottle with less viscous fluid would result in slower decay of oscillations, and vice versa. However, understanding these properties is inherently challenging for autonomous vision systems due to the non-rigid structures and complex behavior of fluids \cite{Tactile_sensing,Visual_perception,percepting_deformables}. In robotics, interactive perception emphasizes the crucial role of physical interaction in perception tasks. It suggests that by physically engaging with the environment, robots can have access to more diverse data and make more informed inferences about their surroundings than they could through passive observation alone \cite{Bohg_2017,Shao_2018}. 
    
\begin{figure*} 
\centering
\includegraphics[width=0.95\textwidth]{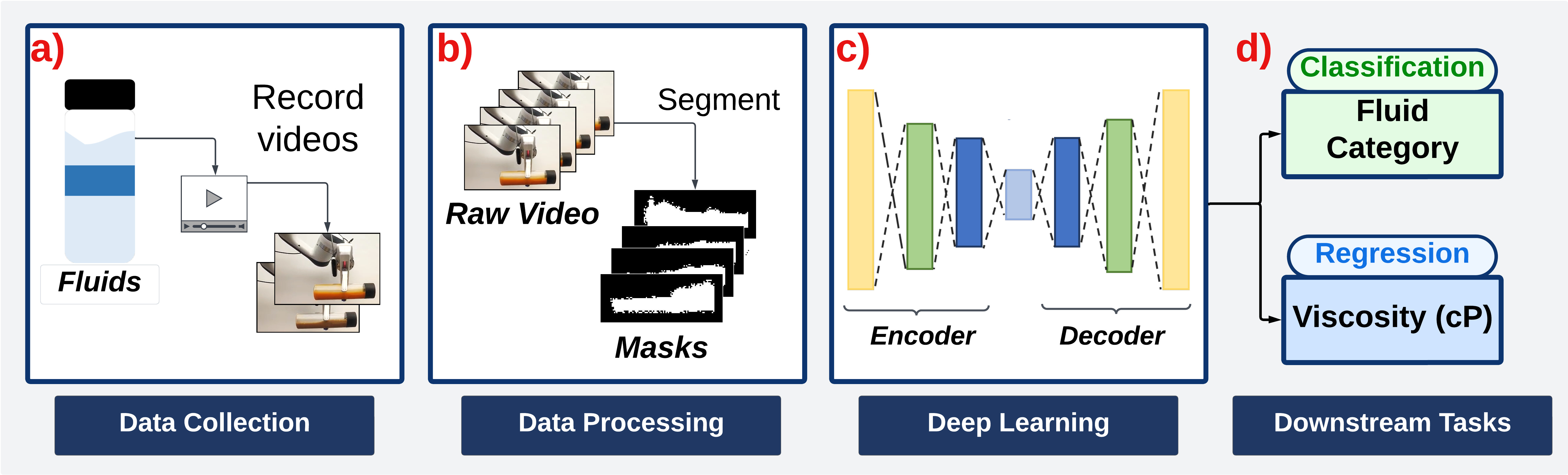}
\caption{\textbf{Overview of the Pipeline.}}
\label{fig:overall}
\end{figure*}

    While a number of approaches such as Reduced-Order Modeling \cite{pant2021deep}, Graph Neural Network-based modeling \cite{li2022graph}, and physics-informed modeling \cite{shu2023physics,phys-inform} have made significant progress in modeling fluid dynamics and accelerating fluid simulations in Computational Fluid Dynamics (CFD), not many work has focused on using data from real-world to make inference on viscosity.
    With the recent increase in machine learning-based fluid modeling  \cite{li2022graph,differentiable_dnn,shu2023physics,vinuesa2022enhancing}, robotic fluid handling \cite{pan2016robot,guevara2017adaptable,kennedy2017precise,kennedy2019autonomous,xian2023fluidlab} and laboratory-based robotic systems \cite{lisca2015towards,burger2020mobile,yoshikawa2022adaptive}, we propose leveraging interactive perception to autonomously estimate viscosity in near-real time.
    
    Our work presents a purely vision-based approach to perform \textbf{two} tasks that can allow the robot to learn the oscillation properties of different fluids from real-world video data: \textbf{1)} classification of fluid types by analyzing videos of five different oscillating fluids and \textbf{2)} prediction of fluid viscosity from videos of oscillating glycerol solutions at eight different viscosity levels. Specifically, we harness the principles of interactive perception by enabling the Franka Emika Panda robot arm to generate oscillations in various fluid samples, and train 3D Convolutional Autoencoder on segmented videos to learn the latent representations through self-supervised learning. Lastly, to perform the two downstream tasks, we discard the decoder and attach the encoder to either a classification or regression head to make the final prediction on fluid category and viscosity respectively.
    The contributions of this paper are:
    \begin{itemize}[leftmargin=*]
        \item We propose an interactive perception approach to build a vision system that non-invasively learns different oscillation patterns in fluids, which could significantly reduce the rigorous effort and time required for experimental viscosity measurement.
        \item Our results demonstrate that the video-based 3D convolutional autoencoder successfully encodes the fluid oscillation patterns into latent representations, which can be transferred for downstream tasks such as fluid class prediction or viscosity estimation.
        \item To the best of our knowledge, this study is the first to investigate the use of video data depicting fluid oscillation patterns for near real-time viscosity inference.
    \end{itemize}

    This paper is organized as follows: Related works are presented in Section \ref{sec:relatedworks}. Section \ref{sec:methods} presents the methodology used to collect the data and train the models as well as details the experimental setup and video dataset collection. Section \ref{sec:results} discusses the findings of the work as well as the results on downstream classification and regression tasks. Lastly, we conclude the paper in Section \ref{sec:conclusion}.

\section{Related Works}
\label{sec:relatedworks}

\textbf{Interactive Perception: }
In robotics, interactive perception (IP) is an active field of research where robots physically interact with their environment to enhance their understanding and perception.\cite{Bohg_2017} Traditional perception methods using passive sensor information are often insufficient to provide robots with a comprehensive understanding of their physical environment. IP can ground the sensor observations and provide the robot with an additional stream of information. Following the advances in machine learning, recent studies have shown promising results on various manipulation and perception tasks employing IP. In an earlier study, K. Hausman et al.\cite{hausman2015active} utilized information from IP with probabilistic models to reduce uncertainties in articulated motion models for different cabinet door configurations. By integrating visual sensing, physical interaction, and intelligent action selection, they demonstrate the robot's ability to effectively handle a variety of articulation models in their task. Work by M. Baum et al.\cite{baum2017opening} utilized active learning in their work on opening a mechanical puzzle called the lockbox, interactively exploring the structures of various lockboxes. In M. A. Lee et al.,\cite{lee2019making} the robot relies on multimodal information such as vision, depth, force-torque, and proprioception obtained through IP to perform robust peg insertion task and demonstrate the transferability of the multimodal latent space. Specifically, they perform self-supervised representation learning on each of multimodal inputs followed by the deep reinforcement learning to find the optimal latent policy. In L. Shao et al.\cite{Shao_2018}, the robot is provided with a deep neural network (DNN) to process the visual effect of its interaction to estimate the scene flow and segmentation masks from two consecutive RGB-D input images. Their IP approach on motion-based object segmentation in a cluttered environment were able to segment well for both the synthetic data and real-world data. More recent work of Shao et al. \cite{shao2020learning} introduces fixtures to the workspace which can be autonomously manipulated by the robot arm. The fixtures provide physical constraints for robot motion and facilitate the reinforcement learning of manipulation skills such as peg insertion, wrench manipulation, and shallow depth insertion.

\textbf{Fluid Modeling and Property Prediction: }
Due to the importance of understanding the behaviors of different fluids and their properties, much work has focused on modeling fluid dynamics and estimating fluid properties. For modeling fluid dynamics, the works of P. Pranshu et al. \cite{pant2021deep} and D. Shu et al. \cite{shu2023physics} both aim to reduce the computational cost of Navier-Stokes equation-based fluid simulations in Computational Fluid Dynamics (CFD). In the works of Z. Li and A. B. Farimani \cite{li2022graph}, they propose Fluid Graph Networks (FGN) to provide a computationally efficient point-based fluid simulator while improving the accuracy and stability of incompressible fluid simulation. In H. J. Huang et al. \cite{Tactile_sensing}, researchers propose a pipeline to predict viscosity through interaction by relating the force measurements when shaking a container of fluid to viscosity. Specifically, they leverage GelSight \cite{yuan2017gelsight}, a visuo-tactile sensor, to sense the force changes caused by the bulk fluid movement, and fit a Gaussian Process Regression model to estimate the liquid properties from these force signals.

\textbf{Segmenting Transparent Fluids: }
Transparent object perception is a challenging task due to their different appearances inherited from the image background \cite{xie2020segmenting}. Despite the difficulties, it is necessary to build systems that can accurately segment transparent objects and fluids to be applied for various fluid-handling robotic/vision systems. J. Liao et al. \cite{liao2020transparent} present an approach to segment transparent objects from video sequences, using trajectory clustering to distinguish the transparent object from the background, however this work focuses on transparent objects and cannot handle transparent liquid within the transparent vessel. In an early work addressing transparent fluid segmentation, A. Yamaguchi and C. G. Atkeson \cite{yamaguchi2016} use optical flow to detect fluid flow as liquid pours from source contained to target container. In the work of C. Schenck and D. Fox \cite{Visual_Closed_loop}, they heat up the liquid and use a thermographic camera to accurately generate ground truth segmentation masks to train a model for the fluid-pouring task. However, this approach is not applicable for our work since ours require data collection at constant temperature to obtain accurate viscosity labels. Another previous approach in segmenting a transparent liquid by G. Narasimhan et al. \cite{narasimhan2022selfsupervised} builds a Generative Adversarial Network (GAN) to convert RGB image of a liquid to transparent ones. This generative model is used to provide synthetic segmentation labels for the transparent liquids. They demonstrate the effectiveness of this vision pipeline by incorporating it into a robotic pouring task, which requires accurate estimation of the fluid quantity through visual inference. However, this system assumes relatively stationary liquid, and is not effective in our case where we consider fluid oscillation as valuable information. S. Eppel et al. \cite{vectorlabpics} presents a model using modern computer vision methodologies to segment and classify medical fluid samples within transparent vessels. They employ a hierarchical, two-layer network to first segment the transparent vessels in the image, and subsequently segment the fluid within the vessels. The model was trained with the Vector-LabPics dataset which consists of 2,187 images of vessels and materials in chemistry lab settings and was used throughout this work for all fluid segmentation steps.

\textbf{Video-Based Autoencoders: } 
There are a number of studies that explores the use of autoencoders trained on video data for a variety of tasks. For instance, K. Wang et al. \cite{videoAE1} proposed Video-Specific Autoencoders, a method that enables the editing and transmission of a single video through latent space arithmetic. In contrast to our work, the Video-Specific Autoencoder consisted of 2D convolutional layers that encode each frame in the video to allow editing of a specific video. However, their use of Principal Component Analysis (PCA) for visualizing what the autoencoder was learning provided valuable insights, and a similar analysis was done in Section \ref{subsec:LatentAnalysis} of our study.
In the work of Z. Lai et al. \cite{videoAE2}, the authors use self-supervised method that utilizes temporal continuity in videos to disentangle and learn representations of 3D structure and camera pose. The autoencoder extracts a temporally-consistent deep voxel feature for the 3D structure and a 3D trajectory for each frame's camera pose, which can be re-entangled for rendering the video frames without the need for ground truth 3D or pose annotations. This method demonstrates its utility in tasks like novel view synthesis, camera pose estimation, and motion-following video generation evaluated on several large-scale natural video datasets.
Another video-based autoencoder approach by Y. Chang et al. \cite{chang2020clustering} employs a unique architecture that learns the spatial and temporal information separately using two autoencoders in parallel for anomaly detection, showing state-of-the-art performance on various video datasets.

\textbf{Real-time Viscosity Inference with Deep Learning: }
For computer vision-based approaches that achieves near real-time viscosity measurement, J. Assen et al. \cite{Visual_perception} train a deep neural network to predict fluid viscosity from image sequences of a simulation-based dataset of fluid splashing, pouring, and stirring. While this method effectively predicts viscosity, relying on simulated data bypasses many of the complexities associated with processing real-world fluid data. 
The work of S. Kv and V. Shenoy \cite{Analysis_Liq_visc} presents a vision-based pipeline for viscosity prediction leveraging the principle of dispersion of incident light. This is also an interaction-free technique to estimate viscosity from the direct relationship between the dispersion of light and the fluid viscosity. They hand-design features to collect from the refracted images of different fluids and train a neural network to predict the viscosity from these features. Despite its effectiveness in real-world data, it requires a very specific experimental setup.
In a similar branch of work by M. Delina et al. \cite{viscosity-with-Yolo}, they apply a high-speed detection model, YOLOv3, to detect the ball in a falling-ball viscometer setup and calculate its falling velocity and viscosity from video input. Although this method achieves a promising average training loss of 0.1424 and average precision (AP) of 82.1\%, it does not achieve real-time inference as the model operates on a video after the falling-ball experiment has concluded.
\begin{figure*}[!ht]
\centering
\includegraphics[width=0.95\textwidth]{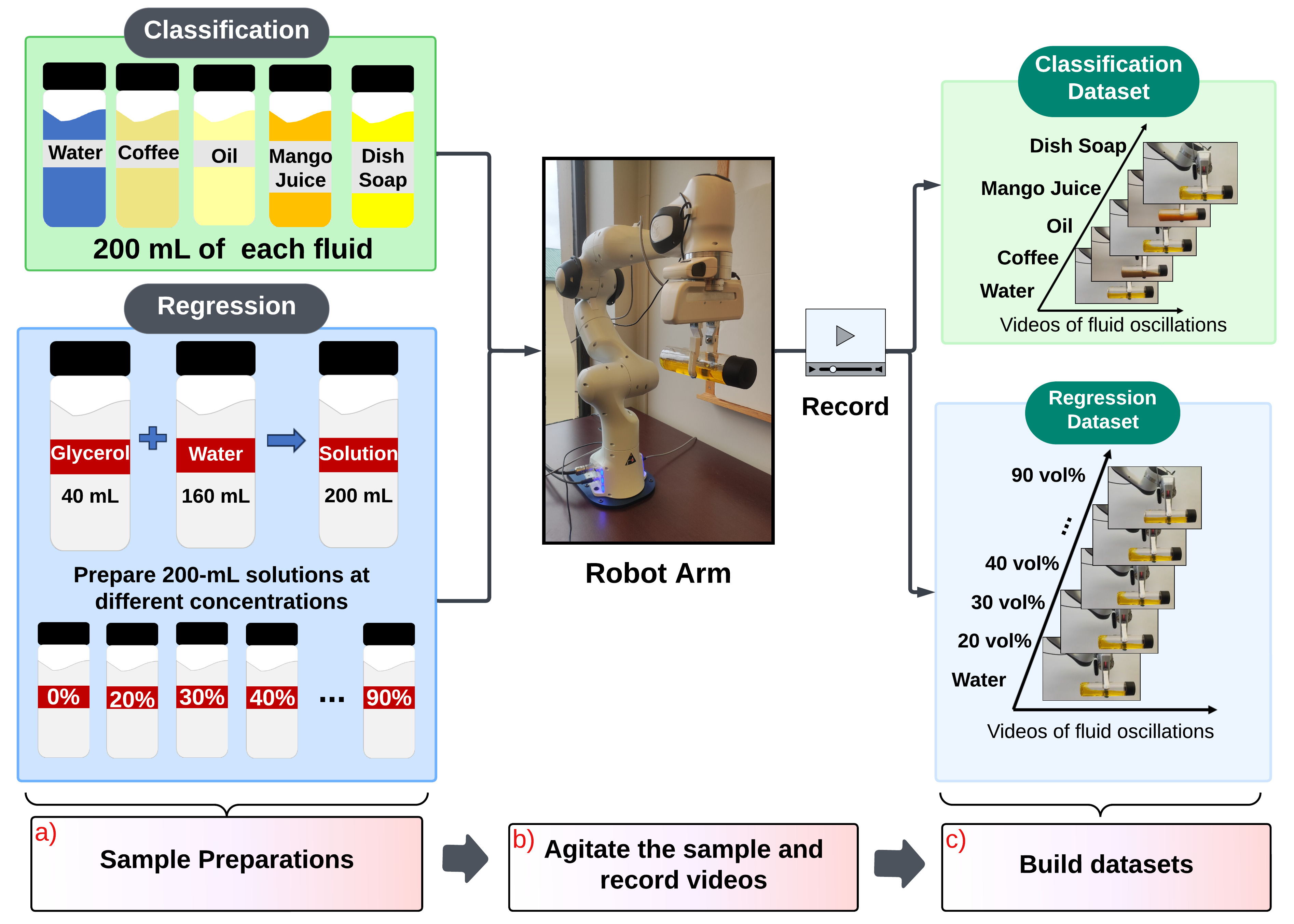}
\caption{\textbf{Interactive Fluid Video Collection.}}
\label{fig:collection}
\end{figure*}

\section{Methods}
\label{sec:methods}
Our pipeline follows four steps as shown in Figure \ref{fig:overall}. We first discuss our experimental setup and the video collection process in Section \ref{subsec:setup}. Next, we discuss the preprocessing steps applied to build the video datasets in Section \ref{subsec:data}. In Section \ref{subsec:autoencoder}, we provide our procedures for training the 3D convolutional autoencoder with oscillation data. Lastly, in Section \ref{subsec:transfer learning}, we briefly discuss the respective methods applied for the downstream classification and regression tasks harnessing the pretrained autoencoder.


\begin{figure*}
\centering
\includegraphics[width=0.9\textwidth]{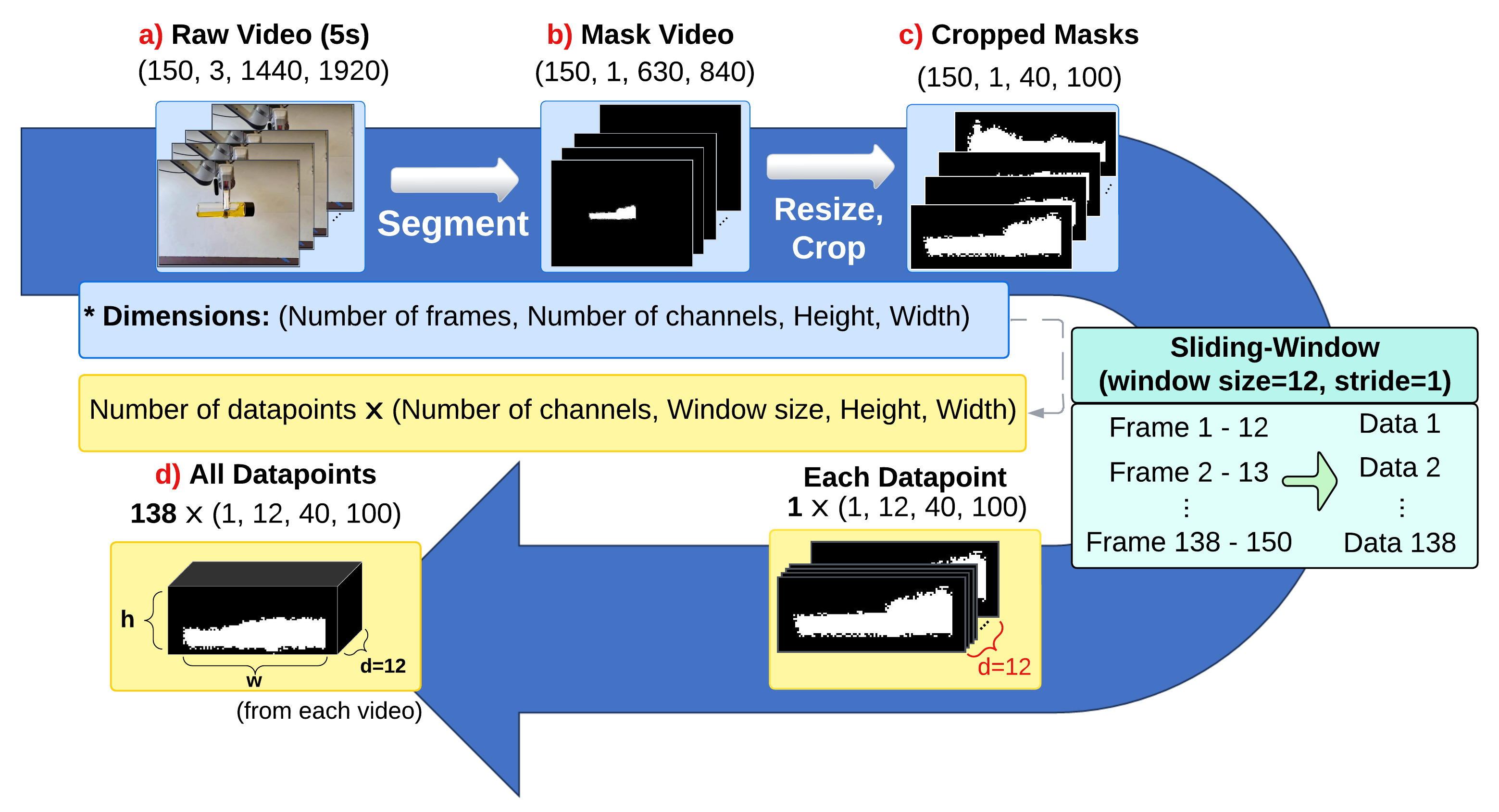}
\caption{\textbf{Video Preprocessing on a Single Video.}}
\label{fig:processing}
\end{figure*}
\subsection{Experimental Setup and Video Collection}
\label{subsec:setup}
    Our experimental setup primarily involves a 7 DoF Franka Emika Panda robot arm, equipped with a gripper to hold and agitate the fluid container. We use cylindrical bottles of identical dimensions (20-cm length and 5-cm diameter) to hold each 200-mL fluid sample at 20 $\degree$C (Figure \ref{fig:collection}-a).
    To facilitate fluid semantic segmentation in Section \ref{subsec:data}, we use a white cardboard background and dye transparent fluid samples, such as water and glycerol solutions, the same color to ensure similar quality in the segmentation results. Video recordings are carried out using a Samsung smartphone camera with a fixed mount, recording at a frame rate of 30 fps and resolution of $(H, W) = (1440, 1920)$.
    
    We control the robot arm with \textit{Frankapy}, a Python wrapper for controlling the Franka robot arm \cite{zhang2020modular}. We restrict the robot's movement to linear shaking motion to yield simpler fluid oscillation patterns. Once the sample is positioned for interaction (Figure \ref{fig:collection}-b), the robot exerts a constant grasping force of 60 N and applies a linear, forward push with a displacement of 30 cm. This movement is conducted at an angle that minimize the visibility of the gripper in the video recordings, allowing the fluid to be completely visible in the recordings. To ensure that any observed oscillations are attributable solely to the fluid's properties and not to handling variability, we maintain consistent force, arm trajectory, and bottle position throughout all recordings.
\begin{table}[H]
\vspace{-2pt}
\centering
\begin{tabularx}{0.95\columnwidth}{X X} 
\hline
\textbf{Fluid Type} & \textbf{Viscosity (cP)} \\
\hline
Water & 1.0  \\
Coffee & 1 - 10 \\
Oil & 25 - 30 \\
Mango Juice & 55 - 75 \\
Dish Soap & 300 - 1000 \\
\hline
\end{tabularx}
\caption{Viscosity ranges of fluids in classification dataset in centipoise (cP) at 20 \degree C}
\label{table:clf_viscosities}
\end{table}
\vspace{-10pt}
    For building the classification dataset (Top of Figure \ref{fig:collection}-c), we prepare five different fluids to record the videos: coffee, dish soap, oil, mango juice, and water. These fluids are not only common in households, but also exhibit significantly different viscosity range (Table \ref{table:clf_viscosities}) which provide a diverse set of oscillation features, from low to high viscosity. Similarly, for the regression dataset (Bottom of Figure \ref{fig:collection}-c), we prepare eight glycerol solutions with varying concentrations (0, 20, 30, 40, 50, 60, 70, 90 vol\%) for extensive range of viscosity for training data. We choose glycerol solution for building the regression dataset due to its wide viscosity range, from 1.0 (pure water) to about 250 centipoise (90 vol\%) at temperature of 20 $\degree$C \cite{visc_lb}, which is significantly broader than that of comparable sucrose solutions.

\subsection{Data Preprocessing}
\label{subsec:data}
    
    We establish separate video datasets for classification and regression tasks using the recordings collected in \ref{subsec:setup}. The classification dataset is composed of 148 videos, with each of the five classification samples contributing around 30 videos. Similarly, the regression dataset consists of 232 videos, with each of the glycerol solution samples contributing approximately 30 videos. To isolate the fluid oscillations from other detectable features and noisy motions in the raw videos, we trim the videos at points where the robot arm remains stationary after it has completed its shaking motion. We also ensure that all trimmed videos within both datasets maintain a consistent video duration of 5 seconds, or 150 frames at 30 fps. The dimension of each video is represented as $(D \times C \times H \times W) = (150 \times 3 \times 1440 \times 1920)$, where $D$ is the number of frames in a video, $C$ is the number of channels, $H$ is the height, and $W$ is the width (Figure \ref{fig:processing}-a).
        
\begin{figure*}[ht]
\centering
\includegraphics[width=0.98\textwidth]{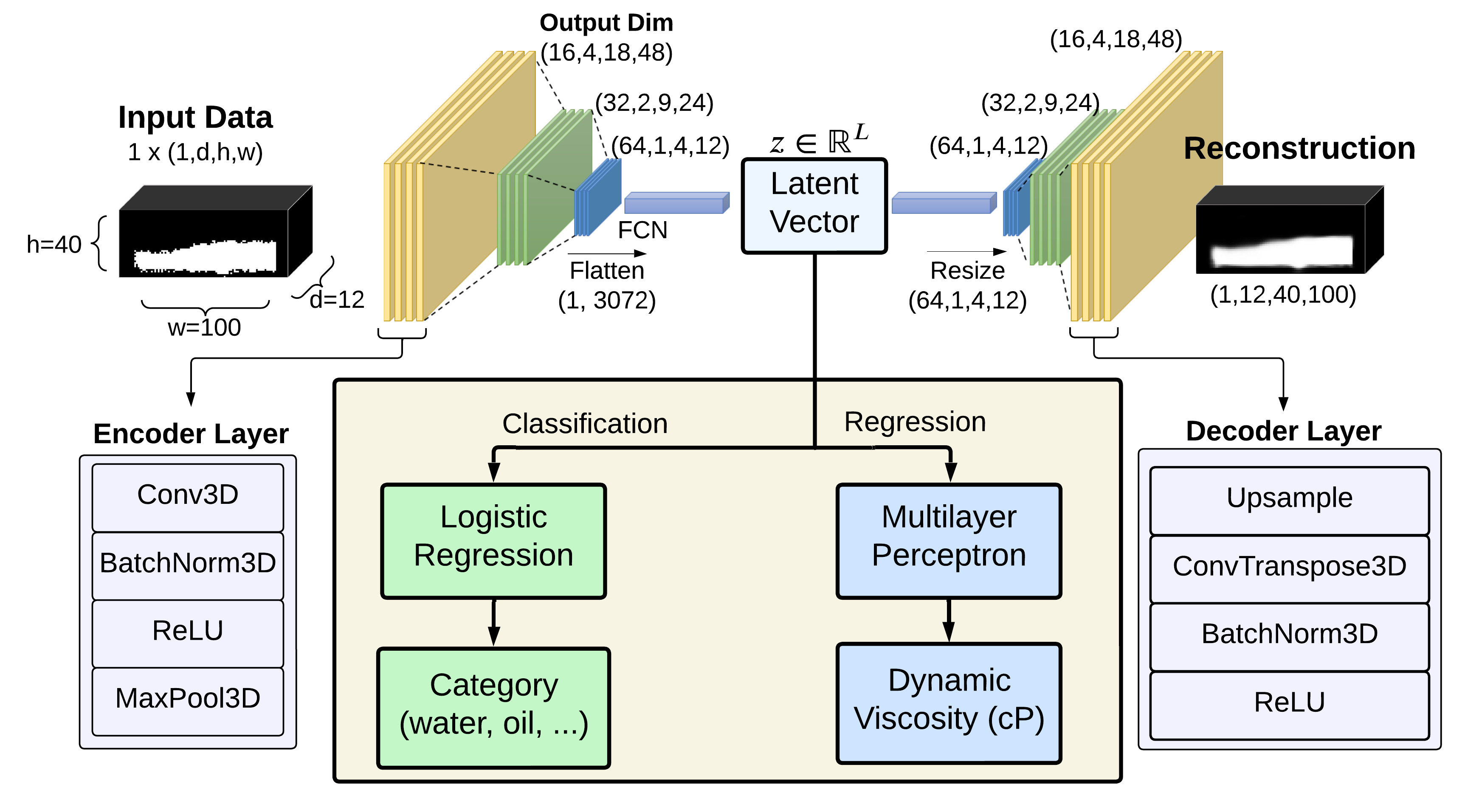}
\caption{\textbf{Autoencoder Architecture and the Downstream Tasks.}}
\label{fig:ae}
\end{figure*}
    To process each of the videos of dimension $(150 \times 3 \times 1440 \times 1920)$ , we mask the fluid portion in it using the semantic segmentation model in S. Eppel et al. \cite{vectorlabpics}, which is a standard fully convolutional neural network trained through parsing the image into regions based on their class. The model reduces the frame size of the RGB input videos from $(H, W) = (1440, 1920)$ to $(630, 840)$ during inference to achieve better segmentation outputs, and number of channels $C$ is reduced to 1 since the videos are segmented as shown in Figure \ref{fig:processing}-b. This masking process eliminates unnecessary information, such as the RGB background or the robot arm, from the raw video and allows the spatio-temporal oscillation of fluid to be the primary focus in the masked videos. Each masked frame is a binary, single-channel image with a pixel value of 1 indicating the fluid region (illustrated as white regions in Figure \ref{fig:processing}).

    To optimize for computational efficiency, each mask video with dimensions $(150 \times 1 \times 630 \times 840)$ is further resized and the fluid regions are cropped, resulting in a reduced size of $(D \times C \times h \times w) = (150 \times 1 \times 40 \times 100)$, where $h$ and $w$ represent the height and width of the cropped frames. Instead of generating non-overlapping splits of mask videos, we apply the sliding-window method with a window size of $d = 12$ frames, and stride of 1 over the first dimension to generate $(D - d)$ datapoints per video. Throughout this work, $D$, $d$, and stride are set to constants of 150, 12, and 1 respectively. Thus, each video is converted into datapoints of dimensions $((D - d) \times C \times d \times h \times w) = (138 \times 1 \times 12 \times 40 \times 100)$, where each of the $(D-d)$ datapoints consists of 12-frame clips from an original 150-frame mask video (Figure \ref{fig:processing}-d). Since this approach generates $(D-d)$, or 138, datapoints for each video, we are able to create a much denser training dataset compared to treating each of the frames as a single datapoint. This method of generating a substantial number of datapoints from each video is especially advantageous in our study, given the time-consuming process of physically collecting video data; each cycle of agitation and recording requires about 30-40 seconds, which makes the collection of a large number of videos impractical.

    After processing, the classification dataset consists of 20,424 datapoints derived through a sliding-window technique from 148 videos, and the regression dataset includes 32,016 datapoints extracted from 232 videos, with each datapoint having the dimension $(C \times d \times h \times w)=(1 \times 12 \times 40 \times 100)$. For the classification dataset, the 148 videos were divided into sets of 118 and 30 videos to generate 16,284 train and 4,140 test classification data respectively. Similarly, for the regression dataset, the 232 videos were split into sets of 200 and 32 videos to form 27,600 train and 4,416 test regression data. Each type of sample in both training datasets is evenly represented with approximately 25 videos. To ensure the robustness of our model, we employed a cross-validation approach, where the videos were randomly split into the aforementioned training and test sets, maintaining the same 118:30 and 200:32 ratio for classification and regression respectively.

\subsection{Autoencoder}
\label{subsec:autoencoder}
    Autoencoders (AEs) \cite{autoencoder,representation_learning} are a type of artificial neural network used to learn efficient latent representations of input data in a self-supervised manner. These latent representations are compact, encoded versions of the input data, and the learning process of an AE is based on minimizing the reconstruction error between the input data (which serves as the ground truth label) and the model's reconstructed output \cite{representation_learning}. In the context of this work, we utilize the 3D convolutional AE, which can perform automatic feature detection from video inputs and learn the spatio-temporal information by leveraging 3D convolutional layers \cite{3dCAE1,3dCAE2}. The latent representations generated by the encoder can be conveniently used for a variety of downstream tasks, specifically, fluid classification and viscosity regression.
    
    $\textbf{Encoder. }$ The encoder, $E(\cdot)$, is composed of three convolutional layers, each of which includes a block of 3D convolution, batch normalization \cite{bn}, ReLU activation \cite{relu}, and max pooling operations with a $2 \times 2 \times 2$ kernel (left part of Figure \ref{fig:ae}). The first convolutional layer uses a $5 \times 5 \times 5$ kernel while the subsequent two layers both use $3 \times 3 \times 3$ kernels. The number of kernels for each layers are 16, 32, and 64 respectively and the stride is constantly set to 1. The layers encode each input sliding-window data of dimensions ($1 \times 12 \times 40 \times 100$), or $I^{C \times d \times h \times w}$, into a reduced 3D representation of dimensions ($64 \times 1 \times 4 \times 12$). Following the convolution layers, an additional fully-connected network (FCN) within the encoder reduces the dimensionality of the flattened 3D representations, ($1 \times 3072$), to a latent vector, $z^{1 \times L}$, where $L$ is the latent dimension.

    $\textbf{Decoder. }$ The decoder, $D(\cdot)$  begins with a FCN to expand the dimension of the latent vector, $z^{1 \times L}$, back to a flattened 3D representation of $1 \times 3072$. Then it is unflattened back to 3D representations of dimension ($64 \times 1 \times 4 \times 12$) to pass into subsequent convolutional layers. As denoted in the right side of Figure \ref{fig:ae}, each decoder convolutional layer includes upsampling layers with trilinear interpolation and transposed convolution followed by batch normalization and ReLU activation to generate a reconstruction of the input, $I'$, that has the same dimension as the input data, $I^{C \times d \times h \times w}$. To ensure the output pixel values fall within the range of 0 to 1, a sigmoid activation function was employed in the output layer, replacing the ReLU activation. Each of the three convolutional layers in the decoder uses 64 $(5 \times 5 \times 5)$ , 32 $(3 \times 3 \times 3)$, and 16 $(3 \times 3 \times 3)$ kernels respectively. This arrangement mirrors the encoder's structure and maintains a symmetric architecture within the autoencoder.
    
    \begin{equation}
    \label{eq:ae1}
    \text{Encoder: } E(I^{C \times d \times h \times w}) = z^{1 \times L}
    \end{equation}

    \begin{equation}
    \label{eq:ae2}
    \text{Decoder: } D(z^{1 \times L}) = I'
    \end{equation}

    We pretrain a symmetric 3D convolutional AE by reconstructing the mask data and minimizing the mean-squared reconstruction loss (MSE) on the train dataset (Eq.\ref{eq:mse}). During training, we augment the input data with subtle translation and rotation transformations to enhance the model's robustness to minor variations in the position and angle of the fluid masks caused by in-hand slippage or shaking of the workspace desk. We allow translations in the range of -2 to 3 pixels and rotations between -2 and 3 degrees for randomly augmenting the input mask videos. We pretrain the autoencoder for 300 epochs and use the test dataset to validate the model after every training epochs, then select the autoencoder with lowest test MSE. To train each of the final classification and regression models, we use a latent dimension $L = 512$, and a batch size of 512. The Adam optimizer is used with a learning rate of 5e-5 for classification and 7e-5 for regression dataset and a weight decay of 1e-5 for both training instances to optimize the objective in Eq.\ref{eq:optim}. The optimal hyperparameters for the AE pretraining were determined by evaluating different combinations of latent dimensions, batch sizes, and learning rates, based on the reconstruction loss observed on the test dataset.
    
    \begin{equation}
    \label{eq:mse}
        L_{recon}(I, I') = \frac{1}{n} \sum_{i=1}^{n} ||I_i - I'_i||^2
    \end{equation}
    
    \begin{equation}
    \label{eq:optim}
        \min_{E(\cdot), \, D(\cdot)} L_{recon}(I, I')
    \end{equation}

\subsection{Transfer Learning for Downstream Tasks}
\label{subsec:transfer learning}
    After training the AE on reconstructing the oscillation patterns, we retain only the encoder, discarding the decoder. This approach utilizes the encoder's ability to transform input data of shape $(C \times d \times h \times w) = (1 \times 12 \times 40 \times 100)$ into latent vectors, $z \in \mathbb{R}^L$.
    For the classification task involving five categories of fluids, we apply multinomial logistic regression on sets of latent vectors. A classifier is fit with all 16,284 train latent vectors to provide a logistic regression model that outputs one-hot encoded class labels. Since each video is converted into multiple datapoints ($D-d$ for each video), we select the class that appears most frequently among all predictions to make a final class prediction at the video level.
    For the viscosity regression task using the glycerol solutions, we attach an additional FCN as a regression head and perform supervised learning with known viscosity labels obtained from \cite{visc_lb}. The network is trained for 200 epochs with a learning rate of 3e-5 and a batch size of 512, using the Adam optimizer to minimize the MSE loss between the predicted and actual viscosity values. To derive a final viscosity prediction at the video level, we average all viscosity predictions from each video. As mentioned in Section \ref{subsec:autoencoder}, we determined the optimal hyperparameters by experimenting with various combinations of latent dimensions, batch sizes, and learning rates.
\begin{figure}[H]
\centering
\includegraphics[width=0.95\columnwidth]{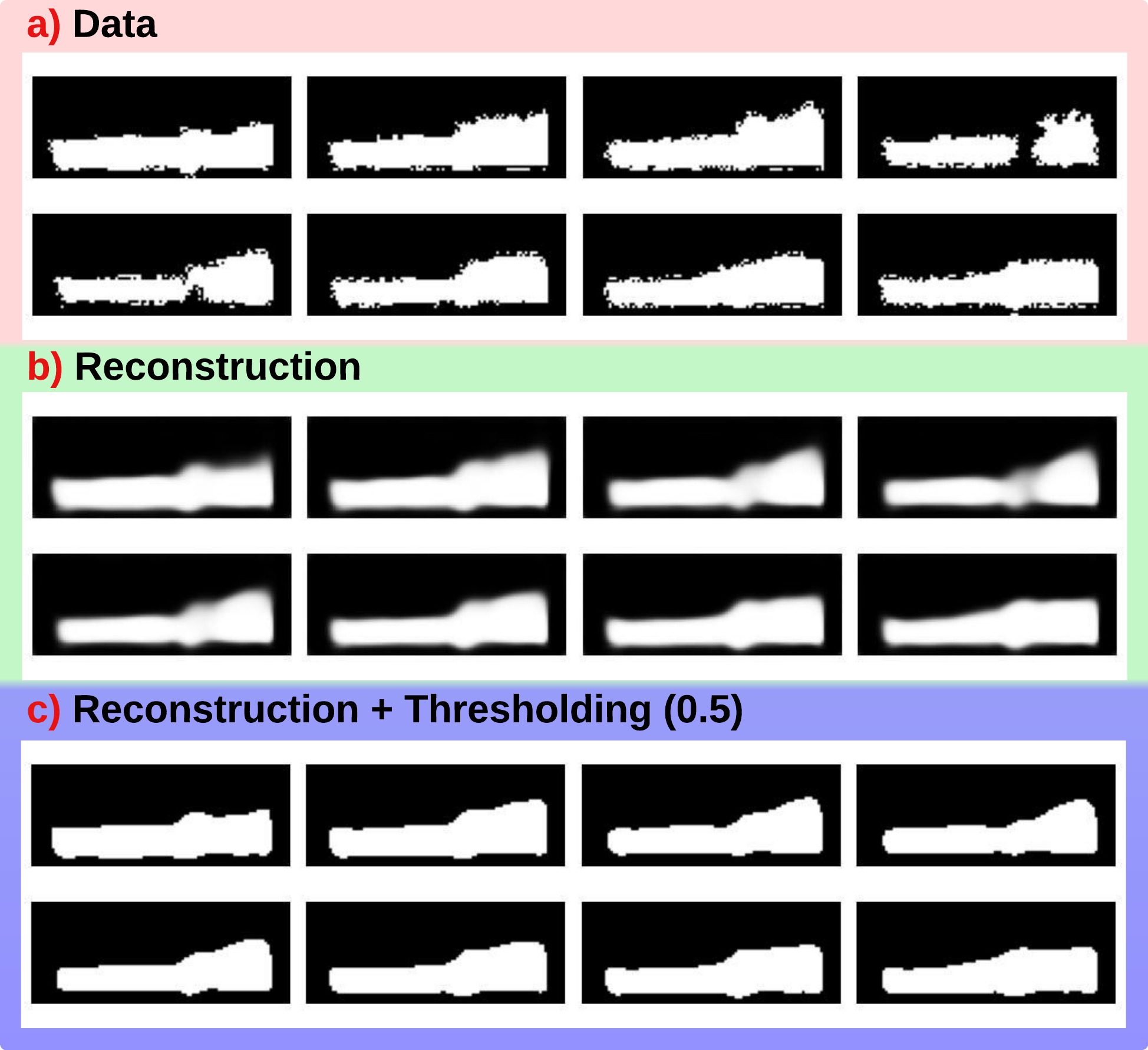}
\caption{\textbf{Segmented Masks and Autoencoder Reconstructions.}}
\label{fig:recon}
\end{figure}

\begin{figure*}[htb]
  \centering
  \begin{subfigure}[b]{0.43\textwidth}
    \includegraphics[width=\textwidth]{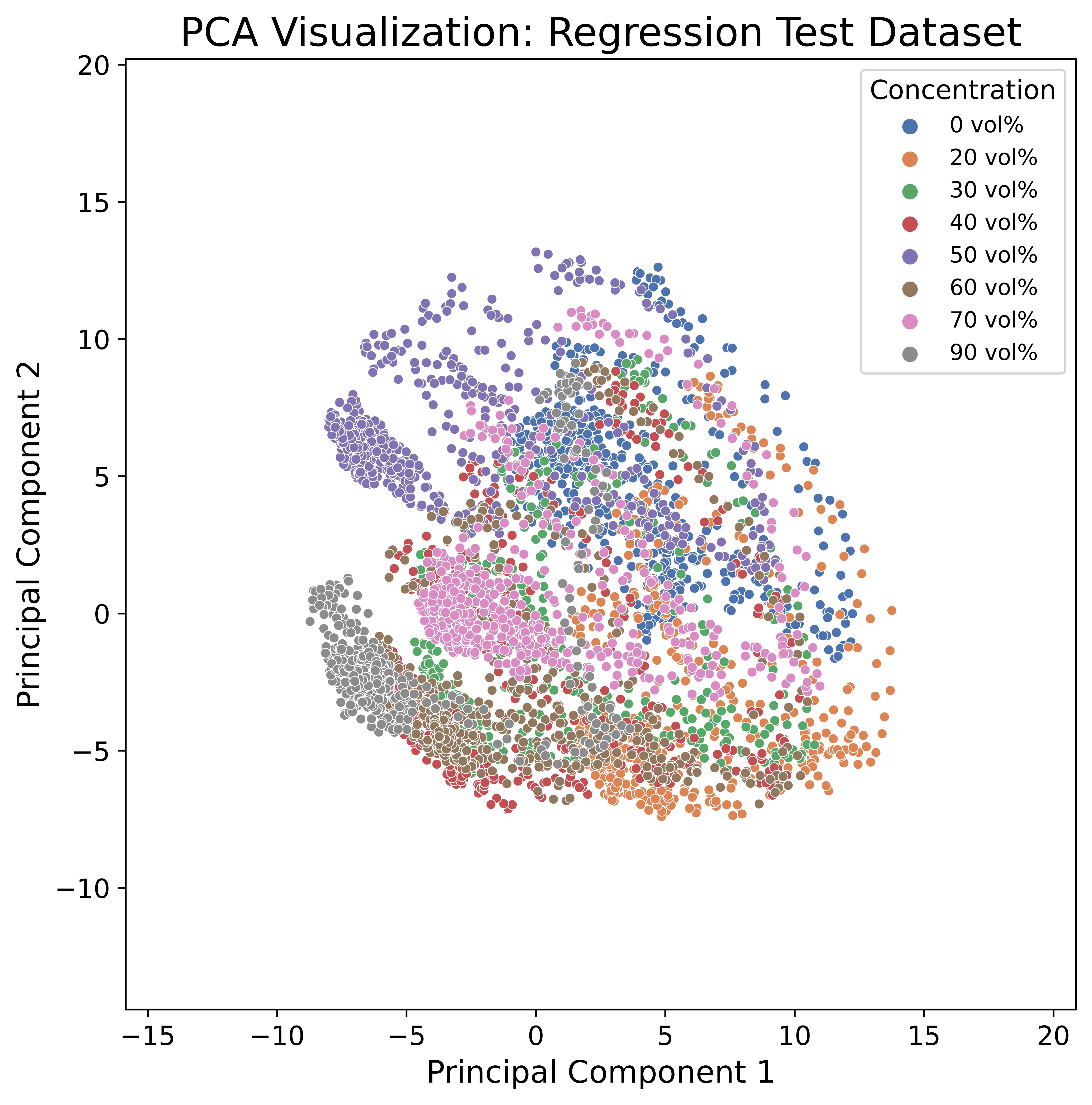}
    \caption{Entire Regression Testset}
    \label{fig:pca_a}
  \end{subfigure}
  \begin{subfigure}[b]{0.5\textwidth}
    \includegraphics[width=\textwidth]{6b_pca_60vol.jpg}
    \caption{60 vol\% Solution}
    \label{fig:pca_b}
  \end{subfigure}
    \caption{\textbf{PCA Projection of Regression Dataset in 2D Space}. (a) Visualizes the latent space of Test Regression Dataset obtained by applying PCA reduction on the latent vectors. (b) Shows the temporal order of frames in a single oscillation video within the projected latent space. All oscillation videos follow a similar layout trend in the latent space.}
  \label{fig:pca}
\end{figure*}
\section{Results}
\label{sec:results}

\subsection{Segmentation Quality}
\label{subsec:segqual}

The mask videos obtained using the pretrained semantic segmentation model from \cite{vectorlabpics} successfully capture the frame level oscillations of fluid as shown in Figure \ref{fig:recon}-a. To use these masks as visual features for differentiating fluid categories or predicting the viscosity of the fluid, it is important to ensure consistent mask quality across different fluid types as the model predictions should be based solely on the oscillation patterns of the fluid, not on other factors such as the shape of the segmented masks, their positions, or any artifacts created by the segmentation process. Inconsistent or imperfect mask qualities can further cause the model to overfit to the masking discrepancies rather than learning the oscillation patterns.
For example, segmented masks on fluids are sometimes discontinuous, as demonstrated in Figure \ref{fig:recon}-a, due to misclassification by the model, which might confuse the parts of the robot's gripper with the fluid sample.

Another challenge in providing consistent masks is in the nature of the Vector Lab-Pics training data used to provide the pretrained segmentation model \cite{vectorlabpics}. The model was primarily trained on images of transparent fluids and solid materials, which can lead to poor segmentation performance when applied to opaque fluids such as mango juice and coffee.
Inconsistent or imperfect mask qualities can lead the model to overfit to discrepancies in masking, rather than learning based on the oscillation patterns. To address this problem, we attempt to collect the videos under strict lighting conditions and maintain a consistent camera angle throughout the video collection process. While the quality of the mask is highly dependent on the camera angle, it remains moderately robust to variations in lighting conditions, which does not significantly impact the generation of segmented videos.

The frames reconstructed by the Autoencoder (AE), as shown in Figure \ref{fig:recon}-b, demonstrate its capability to reconstruct the data frames, capturing the shape and height of the fluid from the original data. The observed blurriness in the reconstruction is due to the use of floating-point numbers resulting from the sigmoid activation function. Figure \ref{fig:recon}-c presents the thresholded reconstruction; however, as thresholding can result in loss of information, we calculate the reconstruction loss (Eq.\ref{eq:mse}) by comparing the original data frames (Figure \ref{fig:recon}-a) with their reconstructed counterparts (Figure \ref{fig:recon}-b) during the AE training.

\subsection{Latent Space Analysis}
\label{subsec:LatentAnalysis}
To gain a deeper understanding of our autoencoder's learning dynamics, we utilize PCA to reduce the dimensionality of the latent representations ($L=512$) and visualize them in 2D plane. This allows us to visually inspect the distribution and relationships between the encoded datapoints, or latent vectors.
A previous work \cite{videoAE1} demonstrates the capability of Video-Specific Autoencoders to learn a temporally continuous representations without any explicit temporal information. Similarly, in this work, we note that each encoded video forms a continuous trajectory within the 2D PCA space even without explicit temporal information. This is exemplified in Figure \ref{fig:pca_b}, where we highlight the trajectory of a 60 vol\% glycerol sample within the entire test dataset (Figure \ref{fig:pca_a} for entire testset).
Furthermore, we observe that the trajectory from different videos exhibit a similar direction and shape, often resembling '\textit{U}' or '\textit{W}' shape directed from right to left as shown in Figure \ref{fig:pca_b}, potentially because the videos were taken under a strict laboratory setup.
Given that video autoencoders can learn the repetitive temporal motion present in a video \cite{videoAE1}, we find that datapoints consisting of earlier frames with more intense oscillation patterns (Light blue 'X's in Figure \ref{fig:pca_b}) are located close to each other in the 2D PCA Space. Conversely, datapoints containing the later oscillations in the video (Magenta 'X's in Figure \ref{fig:pca_b}), where significant decay of oscillation has taken place, are clustered together but located far from the earlier datapoints.
Lastly, when the training of the autoencoder is augmented with translation and rotation (See Section \ref{subsec:autoencoder}), the trajectories form relatively more complex curves, similar to formation of \textit{distinct modes} in \cite{videoAE1}, compared to training without any augmentation.
\begin{figure*}[htb]
  \centering
  \begin{subfigure}[b]{0.48\textwidth}
    \includegraphics[width=\textwidth]{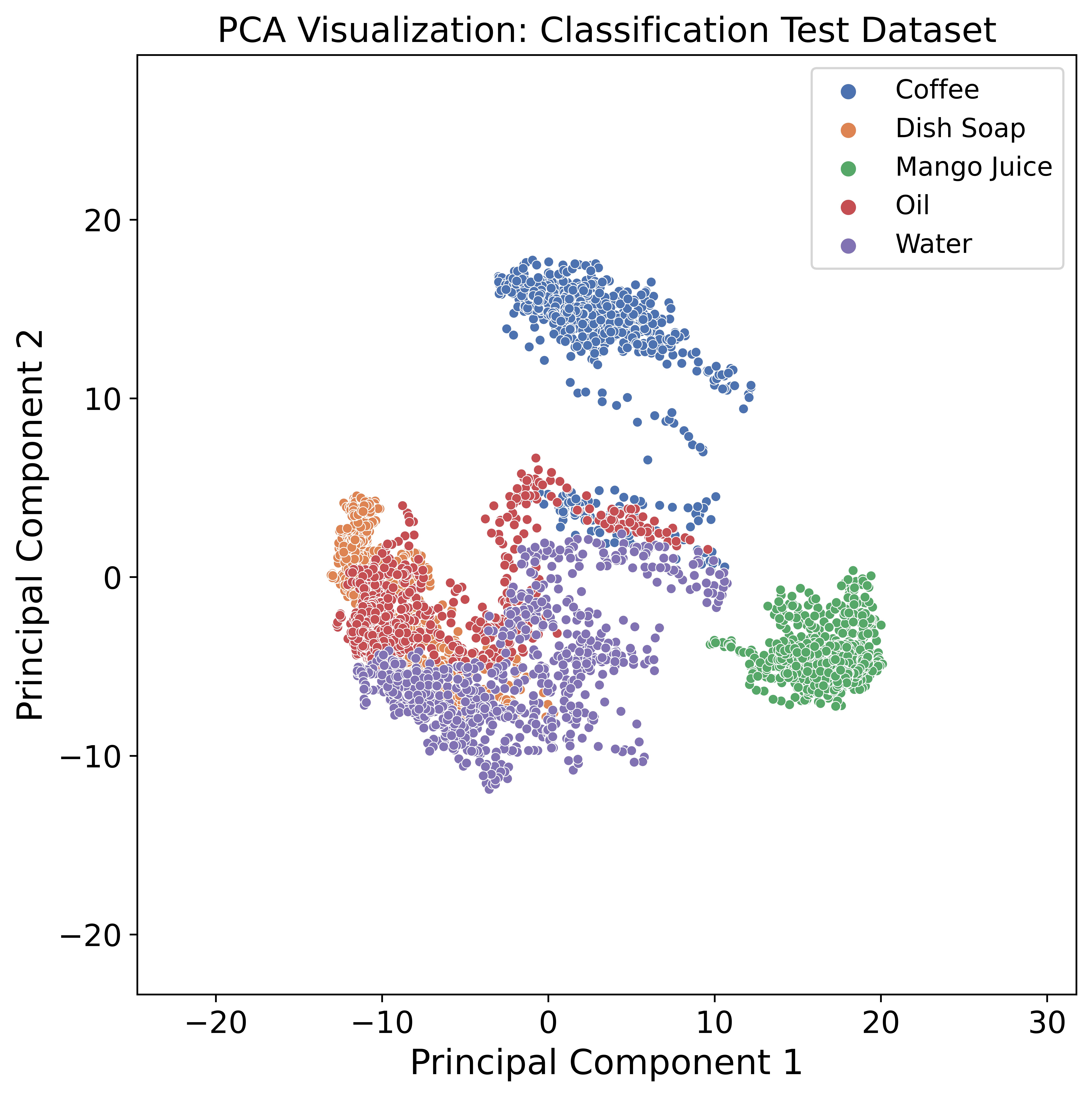}
    \caption{Entire Classification Testset}
    \label{fig:clf_a}
  \end{subfigure}
  \hfill
  \begin{subfigure}[b]{0.48\textwidth}
    \includegraphics[width=\textwidth]{7b_clf_cm.jpg}
    \caption{Confusion Matrix}
    \label{fig:clf_b}
  \end{subfigure}
  \caption{\textbf{PCA Visualization and Classification Results}}
\end{figure*}

\subsection{Classification}
\label{subsec:classification}

In this section, we present our results on classifying fluid categories present in each of the videos in the classification test dataset. We apply multinomial logistic regression to classify the latent vectors encoded with the pretrained autoencoder. In the 2D visualization of the latent representations (Figure \ref{fig:clf_a}), each fluid class forms a distinct cluster. This observation suggests that the latent vectors from the encoder demonstrate consistent characteristics within each class and thus provide a suitable basis for the classification task.

The test dataset for the classification task, derived from the train-test split detailed in Section \ref{subsec:data}, consists of 4,140 datapoints, evenly distributed with 828 datapoints (or 6 videos) for each of the five classes (coffee, dish soap, mango juice, oil, water). Confusion matrix in Figure \ref{fig:clf_b} shows the result of applying logistic regression to the latent vectors generated by the encoder. The classification model (Encoder and Logistic Regression) achieves an data-level accuracy of \textbf{97.1\%} (Eq.\ref{eqn:acc_data}), resulting in 118 misclassifications out of the 4,140 datapoints. These incorrect classifications are from a specific video of the coffee class, potentially due to the quality of the segmentation masks (Section \ref{subsec:segqual}).


\begin{equation}
\label{eqn:acc_data}
\resizebox{.9\columnwidth}{!}{ 
    $\text{Data-level Acc.} = \left( \frac{\text{Correct Data Predictions}}{\text{Number of All Datapoints}} \right) \times 100\%$
}
\end{equation}

\begin{equation}
\label{eqn:acc_video}
\resizebox{.9\columnwidth}{!}{ 
    $\text{Video-level Acc.} = \left( \frac{\text{Correct Video Predictions}}{\text{Number of All Videos}} \right) \times 100\%$
}
\end{equation}

To make class predictions at the video level, we allow the model classify all $D-d=138$ datapoint from a single video and count the greatest number of classes. On a video level, the model achieves accuracy of \textbf{96.6\%} (Eq.\ref{eqn:acc_video}), correctly classifying 29 out of 30 test videos.


\begin{table*}[htb]
\centering
\caption{Comparison of different regression training schemes}
\begin{tabularx}{\textwidth}{l l X X X}
\toprule
& & \multicolumn{3}{c}{Prediction (Mean ± Std)} \\
\cmidrule(lr){3-5}
Volume\% & Label (cP) & Frozen & Unfrozen & Unpretrained \\
\midrule
0 & 1.01 & 1.20 ± 1.57 & 1.00 ± 0.09 & 23.08 ± 22.15 \\
20 & 1.77 & 1.96 ± 1.68 & 1.86 ± 0.12 & 5.44 ± 1.34 \\
30 & 2.50 & 4.58 ± 1.89 & 2.49 ± 0.29 & 9.73 ± 5.28 \\
40 & 3.75 & 5.87 ± 1.60 & 4.21 ± 1.31 & 27.02 ± 27.97 \\
50 & 6.05 & 5.55 ± 1.51 & 5.99 ± 0.10 & 9.63 ± 5.32 \\
60 & 10.96 & 10.02 ± 2.63 & 10.65 ± 0.75 & 21.27 ± 22.44 \\
70 & 22.94 & 23.36 ± 3.01 & 22.86 ± 0.42 & 17.16 ± 14.58 \\
90 & 234.60 & 235.88 ± 4.28 & 234.80 ± 0.41 & 186.74 ± 56.85 \\
\midrule
& MAE & 1.955 & 0.258 & 17.881 \\
\bottomrule
\end{tabularx}
\label{table:reg}
\end{table*}

\subsection{Regression}
\label{subsec:regression}

In our approach, we generate a single viscosity value for each input mask datapoint in the regression test dataset by employing a separate fully-connected network (FCN) that leverages the encoder portion of the AE (Figure \ref{fig:ae}). While there are no existing works directly comparable to our approach of predicting viscosity from visual oscillation patterns of fluid, we conduct an ablation study on the encoder weight to compare different training methodologies. We first train the FCN with the encoder parameters frozen maintaining the pretrained weights. We then train it with the encoder parameters unfrozen, allowing the encoder weights to be fine-tuned along with FCN training. Lastly, we train the entire network for the downstream regression task without pretraining the encoder.

The performance of the regression model (Encoder and the FCN) is assessed based on the mean absolute error computed across all $n=4,416$ label-prediction pairs in the test dataset, denoted as $\nu$ and $\hat{\nu}$ respectively (Eq.\ref{eqn:MAE}). Similar to the classification task, these datapoints are evenly distributed across eight viscosity levels with each level represented by 552 datapoints (i.e., 4,416/8 = 552). For each the eight viscosity levels, we pass the corresponding set of 552 datapoints through the model to obtain viscosity predictions in cP and calculate their mean and standard deviation. These statistical measures presented in Table \ref{table:reg} represent each model's performance across the different viscosity levels.

In Table \ref{table:reg}, it is shown that unfreezing the encoder parameters allows further minimization of the loss and improved performance compared to training with frozen weights as evidenced by the smallest standard deviations and MAE of 0.258. Although the pipeline does not achieve state-of-the-art performance, it shows a promising range of the mean prediction for samples from each viscosity levels, only leveraging the visual oscillation features from videos. The significantly higher MAE (approximately 18 cP) observed for the unpretrained model compared to other pretrained models suggests that pretraining is crucial in enhancing the overall performance of the regression model.

\begin{equation}
\label{eqn:MAE}
\text{MAE} = \frac{1}{n} \sum_{i=1}^{n} |\nu_i - \hat{\nu}_i|
\end{equation}

\section{Conclusion}
\label{sec:conclusion}

In this work, we present a novel interactive perception approach to predicting fluid viscosity purely from visual observation in a time-resolved manner and study the use of segmented masks as feature representing the fluid motion. Our pipeline extracts and learns the segmented masks from videos of oscillating fluids in a self-supervised manner with 3D convolutional autoencoder (AE). Then the pretrained encoder is used to encode the inputs into latent vectors, which can be transferred for downstream tasks such as fluid category classification and viscosity prediction. The final classification pipeline (Encoder and Logistic Regression) yields accuracy of 97.1\%, successfully classifying 4,022 out of 4,140 test datapoints. For the regression task, the pipeline (Encoder and the FCN) with the best performance yields MAE of 0.258 throughout 4,416 test datapoints, which significantly outperforms the scheme without pretraining.

Although we have demonstrated the effectiveness of this method, there are some key limitations that if addressed could result in further improvements, specifically the segmentation mask quality and the generalizability. In this work, we directly leverage the segmentation model from \cite{vectorlabpics}; however, these segmentation masks can sometimes be noisy, introducing more error into the pipeline. Developing a more reliable liquid segmentation module would naturally improve the pipeline as the model training relies on consistent segmentation quality.
While the use of segmented masks contributes to robustness in similar lighting conditions, it also presents a potential limitation in this pipeline's ability to distinguish between different samples that have similar viscosities and oscillation patterns. Furthermore, our methodology involves collecting videos of a consistent, linear shaking motion, which restricts the model's prediction capability on video inputs exhibiting unseen dynamics.

It is plausible that our pipeline could be effective for non-Newtonian fluids provided that sufficient data are available. As a data-driven method, the pipeline has the potential to predict the behavior of non-Newtonian fluids accurately once it is trained with an adequate number of samples. However, it is important to consider the unique property of non-Newtonian fluids, whose viscosity varies with the applied force. This additional complexity makes the task more challenging compared to handling Newtonian fluids and may necessitate the use of a more sophisticated pipeline to accurately capture these dynamics.
A more generalizable extension to this existing work would be to collect a larger dataset encompassing a broader range of fluids, including non-Newtonian fluids, and with variation across lighting conditions, robot sloshing force and motion, etc. to enable model's robust prediction on diverse input data.

Beyond fluid property prediction, we hope to combine this interactive perception module with more complex fluid handling tasks, such that the robot has a container of unidentified fluid, leverages this pipeline to intuit the viscosity, and uses the predicted viscosity to inform the control of a pouring, stirring, or sloshing task.

\section*{Data availability}

All codes used for processing the collected videos and training models are available at GitHub (\url{https://github.com/BaratiLab/vid2visc}). The \texttt{Segmentation} folder contains the modified code for pretrained model based on the previous work of S. Eppel et al. \cite{vectorlabpics} An example of segmented outputs (.avi) for each fluid sample is included in folders \texttt{classification\_dataset} and \texttt{regression\_dataset}. The folder \texttt{autoencoder} contains the python codes for training and the two downstream tasks. All RGB videos and segmented videos used to make the classification and regression dataset can be found in a \href{https://drive.google.com/file/d/1oMa8g1ABoLg6esAilGg66dXpzcmQf1n4/view?usp=drive_link}{public Google Drive} (.zip).

\section*{Acknowledgements}
This work is supported by the Department of Mechanical Engineering, Carnegie Mellon University, United States.
The authors would like to thank our Lab Instructor M. Cline for providing access to the chemistry laboratory facilities and supplying the necessary materials for solution preparation.
\scriptsize{
\bibliographystyle{unsrt}
\bibliography{references}

\begin{thebibliography}{10}

\bibitem{oil_pred}
Xiaodong Gao, Pingchuan Dong, Jiawei Cui, and Qichao Gao.
\newblock Prediction model for the viscosity of heavy oil diluted with light oil using machine learning techniques.
\newblock {\em Energies}, 15(6), 2022.

\bibitem{difficult_viscosity}
Sudhanshu Tiwari, Ajay Dangi, and Rudra Pratap.
\newblock A tip-coupled, two-cantilever, non-resonant microsystem for direct measurement of liquid viscosity.
\newblock {\em Microsystems \& nanoengineering}, 9:34, 03 2023.

\bibitem{viscosity-with-Yolo}
M~Delina, D~S S~P Anugrah, A~M Hussaan, A~F Harlastputra, P~F Akbar, and P~Renaldi.
\newblock Optimizing viscosity measurement: an automated solution with yolov3.
\newblock {\em Journal of Physics: Conference Series}, 2596(1):012022, sep 2023.

\bibitem{pump}
Rouhollah Torabi and Seyyed Nourbakhsh.
\newblock The effect of viscosity on performance of a low specific speed centrifugal pump.
\newblock {\em International Journal of Rotating Machinery}, 2016:1--9, 01 2016.

\bibitem{foodsci}
Arnold~J.{\hspace{0.167em}}T.{\hspace{0.167em}}M. Mathijssen, Maciej Lisicki, Vivek~N. Prakash, and Endre~J.{\hspace{0.167em}}L. Mossige.
\newblock Culinary fluid mechanics and other currents in food science.
\newblock {\em Reviews of Modern Physics}, 95(2), jun 2023.

\bibitem{robotchem1}
Naruki Yoshikawa, Andrew~Zou Li, Kourosh Darvish, Yuchi Zhao, Haoping Xu, Artur Kuramshin, Alán Aspuru-Guzik, Animesh Garg, and Florian Shkurti.
\newblock Chemistry lab automation via constrained task and motion planning, 2023.

\bibitem{robotchem2}
Benjamin Burger, Phillip Maffettone, Vladimir Gusev, Catherine Aitchison, Yang Bai, Xiaoyan Wang, Xiaobo Li, Ben Alston, Buyi Li, Rob Clowes, Nicola Rankin, Brandon Harris, Reiner Sprick, and Andrew Cooper.
\newblock A mobile robotic chemist.
\newblock {\em Nature}, 583:237--241, 07 2020.

\bibitem{robotchem3}
Hatem Fakhruldeen, Gabriella Pizzuto, Jakub Glowacki, and Andrew~Ian Cooper.
\newblock Archemist: Autonomous robotic chemistry system architecture, 2022.

\bibitem{robotchem4}
Aaron Butterworth, Gabriella Pizzuto, Leszek Pecyna, Andrew~I. Cooper, and Shan Luo.
\newblock Leveraging multi-modal sensing for robotic insertion tasks in r \& d laboratories, 2023.

\bibitem{robotchem5}
Parisa Shiri, Veronica Lai, Tara Zepel, Daniel Griffin, Jonathan Reifman, Sean Clark, Shad Grunert, Lars Yunker, Sebastian Steiner, Henry Situ, Fan Yang, Paloma Prieto, and Jason Hein.
\newblock Automated solubility screening platform using computer vision.
\newblock {\em iScience}, 24:102176, 02 2021.

\bibitem{Tactile_sensing}
Hung-Jui Huang, Xiaofeng Guo, and Wenzhen Yuan.
\newblock Understanding dynamic tactile sensing for liquid property estimation.
\newblock {\em arXiv preprint arXiv:2205.08771}, 2022.

\bibitem{viscous_decay}
Michael Sinhuber, Eberhard Bodenschatz, and Gregory~P. Bewley.
\newblock Decay of turbulence at high reynolds numbers.
\newblock {\em Phys. Rev. Lett.}, 114:034501, Jan 2015.

\bibitem{Visual_perception}
Jan Assen, Shin'ya Nishida, and Roland Fleming.
\newblock Visual perception of liquids: Insights from deep neural networks.
\newblock {\em PLOS Computational Biology}, 16:e1008018, 08 2020.

\bibitem{percepting_deformables}
Carolyn Matl.
\newblock {\em Interactive Perception for Robotic Manipulation of Liquids, Grains, and Doughs}.
\newblock PhD thesis, EECS Department, University of California, Berkeley, Aug 2021.

\bibitem{Bohg_2017}
Jeannette Bohg, Karol Hausman, Bharath Sankaran, Oliver Brock, Danica Kragic, Stefan Schaal, and Gaurav~S. Sukhatme.
\newblock Interactive perception: Leveraging action in perception and perception in action.
\newblock {\em {IEEE} Transactions on Robotics}, 33(6):1273--1291, dec 2017.

\bibitem{Shao_2018}
Lin Shao, Parth Shah, Vikranth Dwaracherla, and Jeannette Bohg.
\newblock Motion-based object segmentation based on dense {RGB}-d scene flow.
\newblock {\em {IEEE} Robotics and Automation Letters}, 3(4):3797--3804, oct 2018.

\bibitem{pant2021deep}
Pranshu Pant, Ruchit Doshi, Pranav Bahl, and Amir Barati~Farimani.
\newblock Deep learning for reduced order modelling and efficient temporal evolution of fluid simulations.
\newblock {\em Physics of Fluids}, 33(10), 2021.

\bibitem{li2022graph}
Zijie Li and Amir~Barati Farimani.
\newblock Graph neural network-accelerated lagrangian fluid simulation.
\newblock {\em Computers \& Graphics}, 103:201--211, 2022.

\bibitem{shu2023physics}
Dule Shu, Zijie Li, and Amir~Barati Farimani.
\newblock A physics-informed diffusion model for high-fidelity flow field reconstruction.
\newblock {\em Journal of Computational Physics}, 478:111972, 2023.

\bibitem{phys-inform}
Hamidreza Eivazi and Ricardo Vinuesa.
\newblock Physics-informed deep-learning applications to experimental fluid mechanics, 2022.

\bibitem{differentiable_dnn}
Connor Schenck and Dieter Fox.
\newblock Spnets: Differentiable fluid dynamics for deep neural networks.
\newblock In Aude Billard, Anca Dragan, Jan Peters, and Jun Morimoto, editors, {\em Proceedings of The 2nd Conference on Robot Learning}, volume~87 of {\em Proceedings of Machine Learning Research}, pages 317--335. PMLR, 29--31 Oct 2018.

\bibitem{vinuesa2022enhancing}
Ricardo Vinuesa and Steven~L Brunton.
\newblock Enhancing computational fluid dynamics with machine learning.
\newblock {\em Nature Computational Science}, 2(6):358--366, 2022.

\bibitem{pan2016robot}
Zherong Pan, Chonhyon Park, and Dinesh Manocha.
\newblock Robot motion planning for pouring liquids.
\newblock In {\em Proceedings of the international conference on automated planning and scheduling}, volume~26, pages 518--526, 2016.

\bibitem{guevara2017adaptable}
Tatiana~L{\'o}pez Guevara, Nicholas~Kenelm Taylor, Michael Gutmann, Subramanian Ramamoorthy, and Kartic Subr.
\newblock Adaptable pouring: Teaching robots not to spill using fast but approximate fluid simulation.
\newblock In {\em 1st Conference on Robot Learning 2017}, pages 77--86, 2017.

\bibitem{kennedy2017precise}
Monroe Kennedy, Kendall Queen, Dinesh Thakur, Kostas Daniilidis, and Vijay Kumar.
\newblock Precise dispensing of liquids using visual feedback.
\newblock In {\em 2017 IEEE/RSJ international conference on intelligent robots and systems (IROS)}, pages 1260--1266. IEEE, 2017.

\bibitem{kennedy2019autonomous}
Monroe Kennedy, Karl Schmeckpeper, Dinesh Thakur, Chenfanfu Jiang, Vijay Kumar, and Kostas Daniilidis.
\newblock Autonomous precision pouring from unknown containers.
\newblock {\em IEEE Robotics and Automation Letters}, 4(3):2317--2324, 2019.

\bibitem{xian2023fluidlab}
Zhou Xian, Bo~Zhu, Zhenjia Xu, Hsiao-Yu Tung, Antonio Torralba, Katerina Fragkiadaki, and Chuang Gan.
\newblock Fluidlab: A differentiable environment for benchmarking complex fluid manipulation.
\newblock {\em arXiv preprint arXiv:2303.02346}, 2023.

\bibitem{lisca2015towards}
Gheorghe Lisca, Daniel Nyga, Ferenc B{\'a}lint-Bencz{\'e}di, Hagen Langer, and Michael Beetz.
\newblock Towards robots conducting chemical experiments.
\newblock In {\em 2015 IEEE/RSJ International Conference on Intelligent Robots and Systems (IROS)}, pages 5202--5208. IEEE, 2015.

\bibitem{burger2020mobile}
Benjamin Burger, Phillip~M Maffettone, Vladimir~V Gusev, Catherine~M Aitchison, Yang Bai, Xiaoyan Wang, Xiaobo Li, Ben~M Alston, Buyi Li, Rob Clowes, et~al.
\newblock A mobile robotic chemist.
\newblock {\em Nature}, 583(7815):237--241, 2020.

\bibitem{yoshikawa2022adaptive}
Naruki Yoshikawa, Andrew~Zou Li, Kourosh Darvish, Yuchi Zhao, Haoping Xu, Alan Aspuru-Guzik, Animesh Garg, and Florian Shkurti.
\newblock An adaptive robotics framework for chemistry lab automation.
\newblock {\em arXiv preprint arXiv:2212.09672}, 2022.

\bibitem{hausman2015active}
Karol Hausman, Scott Niekum, Sarah Osentoski, and Gaurav~S Sukhatme.
\newblock Active articulation model estimation through interactive perception.
\newblock In {\em 2015 IEEE International Conference on Robotics and Automation (ICRA)}, pages 3305--3312. IEEE, 2015.

\bibitem{baum2017opening}
Manuel Baum, Matthew Bernstein, Roberto Martin-Martin, Sebastian H{\"o}fer, Johannes Kulick, Marc Toussaint, Alex Kacelnik, and Oliver Brock.
\newblock Opening a lockbox through physical exploration.
\newblock In {\em 2017 IEEE-RAS 17th International Conference on Humanoid Robotics (Humanoids)}, pages 461--467. IEEE, 2017.

\bibitem{lee2019making}
Michelle~A. Lee, Yuke Zhu, Peter Zachares, Matthew Tan, Krishnan Srinivasan, Silvio Savarese, Li~Fei-Fei, Animesh Garg, and Jeannette Bohg.
\newblock Making sense of vision and touch: Learning multimodal representations for contact-rich tasks, 2019.

\bibitem{shao2020learning}
Lin Shao, Toki Migimatsu, and Jeannette Bohg.
\newblock Learning to scaffold the development of robotic manipulation skills, 2020.

\bibitem{yuan2017gelsight}
Wenzhen Yuan, Siyuan Dong, and Edward~H Adelson.
\newblock Gelsight: High-resolution robot tactile sensors for estimating geometry and force.
\newblock {\em Sensors}, 17(12):2762, 2017.

\bibitem{xie2020segmenting}
Enze Xie, Wenjia Wang, Wenhai Wang, Mingyu Ding, Chunhua Shen, and Ping Luo.
\newblock Segmenting transparent objects in the wild.
\newblock In {\em Computer Vision--ECCV 2020: 16th European Conference, Glasgow, UK, August 23--28, 2020, Proceedings, Part XIII 16}, pages 696--711. Springer, 2020.

\bibitem{liao2020transparent}
Jie Liao, Yanping Fu, Qingan Yan, and Chunxia Xiao.
\newblock Transparent object segmentation from casually captured videos.
\newblock {\em Computer Animation and Virtual Worlds}, 31(4-5):e1950, 2020.

\bibitem{yamaguchi2016}
Akihiko Yamaguchi and Christopher~G. Atkeson.
\newblock Stereo vision of liquid and particle flow for robot pouring.
\newblock In {\em 2016 IEEE-RAS 16th International Conference on Humanoid Robots (Humanoids)}, pages 1173--1180, 2016.

\bibitem{Visual_Closed_loop}
Connor Schenck and Dieter Fox.
\newblock Visual closed-loop control for pouring liquids, 2017.

\bibitem{narasimhan2022selfsupervised}
Gautham~Narayan Narasimhan, Kai Zhang, Ben Eisner, Xingyu Lin, and David Held.
\newblock Self-supervised transparent liquid segmentation for robotic pouring, 2022.

\bibitem{vectorlabpics}
Sagi Eppel, Haoping Xu, and Al{\'{a}}n Aspuru{-}Guzik.
\newblock Computer vision for liquid samples in hospitals and medical labs using hierarchical image segmentation and relations prediction.
\newblock {\em CoRR}, abs/2105.01456, 2021.

\bibitem{videoAE1}
Kevin Wang, Deva Ramanan, and Aayush Bansal.
\newblock Video exploration via video-specific autoencoders.
\newblock {\em CoRR}, abs/2103.17261, 2021.

\bibitem{videoAE2}
Zihang Lai, Sifei Liu, Alexei~A. Efros, and Xiaolong Wang.
\newblock Video autoencoder: self-supervised disentanglement of static 3d structure and motion, 2021.

\bibitem{chang2020clustering}
Yunpeng Chang, Zhigang Tu, Wei Xie, and Junsong Yuan.
\newblock Clustering driven deep autoencoder for video anomaly detection.
\newblock In {\em Computer Vision--ECCV 2020: 16th European Conference, Glasgow, UK, August 23--28, 2020, Proceedings, Part XV 16}, pages 329--345. Springer, 2020.

\bibitem{Analysis_Liq_visc}
Santhosh Kv and Vignesh Shenoy.
\newblock Analysis of liquid viscosity by image processing techniques.
\newblock {\em Indian Journal of Science and Technology}, 9, 08 2016.

\bibitem{zhang2020modular}
Kevin Zhang, Mohit Sharma, Jacky Liang, and Oliver Kroemer.
\newblock A modular robotic arm control stack for research: Franka-interface and frankapy.
\newblock {\em arXiv preprint arXiv:2011.02398}, 2020.

\bibitem{visc_lb}
John~Bartlett Segur and Helen~E. Oberstar.
\newblock Viscosity of glycerol and its aqueous solutions.
\newblock {\em Industrial \& Engineering Chemistry}, 43:2117--2120, 1951.

\bibitem{autoencoder}
Ian~J. Goodfellow, Yoshua Bengio, and Aaron Courville.
\newblock {\em Deep Learning}.
\newblock MIT Press, Cambridge, MA, USA, 2016.
\newblock \url{http://www.deeplearningbook.org}.

\bibitem{representation_learning}
Yoshua Bengio, Aaron Courville, and Pascal Vincent.
\newblock Representation learning: A review and new perspectives, 2014.

\bibitem{3dCAE1}
Hiroyuki Yamaguchi, Yuki Hashimoto, Genichi Sugihara, Jun Miyata, Toshiya Murai, Hidehiko Takahashi, Manabu Honda, Akitoyo Hishimoto, and Yuichi Yamashita.
\newblock Three-dimensional convolutional autoencoder extracts features of structural brain images with a “diagnostic label-free” approach: Application to schizophrenia datasets.
\newblock {\em Frontiers in Neuroscience}, 15, 2021.

\bibitem{3dCAE2}
Ali Mjalled, Reza Namdar, Lucas Reineking, Mohammad Norouzi, Fathollah Varnik, and Martin Mönnigmann.
\newblock Parametric 3d convolutional autoencoder for the prediction of flow fields in a bed configuration of hot particles, 2023.

\bibitem{bn}
Sergey Ioffe and Christian Szegedy.
\newblock Batch normalization: Accelerating deep network training by reducing internal covariate shift, 2015.

\bibitem{relu}
Alex Krizhevsky, Ilya Sutskever, and Geoffrey Hinton.
\newblock Imagenet classification with deep convolutional neural networks.
\newblock {\em Neural Information Processing Systems}, 25, 01 2012.

\end{thebibliography}
}

\end{multicols}
\end{document}